# Designing LMS and Instructional Strategies for Integrating Generative-Conversational AI

Elias Ra, Seung Je Kim, Eui-Yeong Seo, Geunju So*

* Corresponding Author

Abstract

Higher education faces growing challenges in delivering personalized, scalable, and pedagogically coherent learning experiences. This study introduces a structured framework for designing an AI-powered Learning Management System (AI-LMS) that integrates generative and conversational AI to support adaptive, interactive, and learner-centered instruction. Using a design-based research (DBR) methodology, the framework unfolds through five phases: literature review, SWOT analysis, development of ethical-pedagogical principles, system design, and instructional strategy formulation. The resulting AI-LMS features modular components—including configurable prompts, adaptive feedback loops, and multi-agent conversation flows—aligned with pedagogical paradigms such as behaviorist, constructivist, and connectivist learning theories. By combining AI capabilities with human-centered design and ethical safeguards, this study advances a practical model for AI integration in education. Future research will validate and refine the system through real-world implementation.

## 1. Introduction

The integration of artificial intelligence (AI) into education is rapidly reshaping how students learn and how educators teach. Among emerging innovations, generative and conversational AI powered by large language models are making a profound impact on educational experiences. These technologies enable personalized learning, interactive dialogue, and efficient content generation, potentially addressing longstanding challenges in student engagement, instructional scalability, and formative assessment. However, effective implementation requires a nuanced understanding of both the technical capabilities and the pedagogical strategies associated with AI in educational contexts.

This study explores the design of an AI-powered learning management system (LMS) that leverages generative-conversational AI to support teaching and learning in higher education. It addresses four central research questions:

- What are the technical characteristics of AI that enhance learning in an educational setting?
- What pedagogical principles should guide the adoption of AI in education to ensure effective learning?
- How can an AI-powered LMS be designed to support both educators and students while maintaining instructional integrity?
- How can strategic applications of AI-powered LMS align with different pedagogical approaches?

2. Methodology

This study adopts a Design-Based Research (DBR) methodology (Brown, 1992; Collins, 1992; Wang & Hannafin, 2005; Reeves, 2006), which is well-suited for investigating complex educational interventions in authentic contexts. DBR emphasizes the iterative design, implementation, and refinement of educational innovations while generating both practical outcomes and theoretical insights. It is particularly appropriate for the development of technology-enhanced learning environments, such as AI-powered learning systems, where instructional design is informed by empirical feedback. This paper presents the initial phase of the DBR process, focusing on the conceptualization and development of an AI-integrated Learning Management System (AI-LMS) that aligns with pedagogical principles, ethical considerations, and instructional strategies.

The research follows a five-phase methodological structure designed to investigate the technical, pedagogical, and instructional dimensions of integrating generative-conversational AI into higher education. Each phase corresponds to a stage in the DBR process and contributes to the design and prototyping of an AI-powered LMS, as well as the development of theory-informed instructional strategies.

Phase 1: Literature Review

This phase involves a comprehensive analysis of recent scholarly literature on artificial intelligence in education. It identifies the technical characteristics of AI technologies—such as natural language processing, generative models, and conversational agents—that support learning, engagement, and instructional effectiveness. This phase establishes the technological and conceptual foundation for the subsequent design process.

Phase 2: SWOT Analysis

In this phase, building on the findings of the literature review, we conduct a critical SWOT analysis to examine the strengths, weaknesses, opportunities, and threats associated with AI integration in education. This analysis explores how AI can advance educational equity, accommodate learner diversity, and enhance instructional quality, while also addressing significant challenges such as ethical dilemmas, technical constraints, and institutional barriers.

Phase 3: Principle Building

Synthesizing insights from the SWOT analysis, this phase formulates a set of guiding pedagogical and ethical principles for AI adoption in education. These principles—such as pedagogical alignment, human-centered design, ethical inclusion, and data privacy—provide a normative foundation for designing and implementing AI-enhanced instructional systems in a responsible and learner-centered manner.

Phase 4: AI-LMS System Design

Building on the technological insights from the literature and the principles derived in the previous phase, this stage focuses on the conceptual and technical design of the AI Learning Mate System (AI-LMS). It outlines system components such as AI role configuration, prompt design, content integration, interactive feedback mechanisms, and collaborative conversation flows to ensure instructional alignment and usability.

Phase 5: Instructional Strategy Development

This final phase involves designing exemplar instructional activities—such as drills, discussions, debates, and collaborative projects—that demonstrate how the AI-LMS can be implemented across varied pedagogical frameworks. These instructional patterns reflect learning theories such as behaviorism, constructivism, and connectivism, and highlight how AI-mediated dialogue can foster personalized, adaptive, and collaborative learning.

The literature review defines the educational problem and technological context (Phase 1). The SWOT analysis evaluates strengths and limitations (Phase 2). A principle-building phase (Phase 3) synthesizes ethical and pedagogical guidelines. The AI-LMS is then conceptualized as a prototype (Phase 4), and instructional strategies are mapped to theoretical models (Phase 5). While this paper presents the design phase, future work will extend the DBR cycle through implementation, empirical evaluation, and iterative refinement in real-world educational settings.

## 3. AI in Education

### 3-1. AI Technology for Education (on RQ1)

The past five years have witnessed a transformative shift in artificial intelligence, driven by generative AI models and conversational systems. These technologies, while only recently adopted in education, have long been explored in broader AI research, industry applications, and human-computer interaction (HCI).

In this section, we first examine foundational literature to establish a baseline understanding of AI technologies, and then classify recent research trends related to their application in educational contexts. Specifically, this review of AI integration in education is grounded in a close analysis of publications from the past three years (2022-2025), drawn from leading peer-reviewed journals in the field of educational technology. It synthesizes emerging trends, key technological innovations, and critical pedagogical and ethical issues that are shaping the use of AI in educational practice.

#### 3-1-1. Foundational Understanding of AI Technologies

Generative AI refers to systems that create new content—text, code, images, or audio—based on patterns learned from vast datasets. Its most visible examples are Large Language Models (LLMs) like OpenAI's GPT-4, Anthropic's Claude, and Google's Gemini, which have dramatically expanded the capacity of machines to simulate human-like reasoning and language fluency.

A foundational work in this space is the "Attention Is All You Need" paper by Vaswani et al. (2017), which introduced the Transformer architecture—a model based entirely on attention mechanisms rather than recurrence. This architecture laid the groundwork for subsequent large language models. Building on this foundation, the GPT-3 paper by Brown et al. (2020) introduced the idea that language models could perform complex tasks with little or no additional training ("few-shot learning"). This was followed by OpenAI's release of GPT-4 (OpenAI, 2023), which demonstrated even greater reasoning abilities and support for multimodal inputs.

Recent general reviews such as Bommasani et al. (2021) coined the term "foundation models" to describe how these systems serve as flexible platforms for a wide range of downstream tasks—from legal writing and coding to healthcare diagnostics and product design. Conversational AI refers to dialogue-based systems—such as chatbots, virtual assistants, or multi-modal agents—that engage users through natural language. While early systems relied on predefined rules, modern agents use LLMs to dynamically generate and sustain conversation, with real-time responsiveness and contextual memory.

In industry, platforms like XiaoIce (Zhou et al., 2020) and Google Assistant have pushed the boundary of emotional engagement, voice interface, and long-form dialogue. In HCI research, scholars have emphasized the importance of trust, persona design, and interactional ethics in shaping how people relate to AI.

Rather than programming with code, users now instruct AI through natural language prompts. This shift has given rise to the field of prompt engineering—the strategic crafting of inputs to elicit useful, accurate, or creative outputs from models. Techniques such as chain-of-thought prompting (Wei et al., 2022) demonstrate that even simple changes in prompt formatting can unlock reasoning capabilities in large language models (LLMs).

As AI adoption broadens, prompt engineering is emerging as a practical skill for non-technical users across domains such as journalism, design, and research. This trend aligns closely with the needs of non-engineer educators—a key consideration in the AI-LMS framework proposed in this study. In particular, chain-of-thought prompting, inspired by human cognitive processes, is grounded in principles of metacognition and educational psychology. In this sense, conversational AI can be purposefully employed for educational purposes when supported by conversation design embedded within well-structured instructional design frameworks. This potential will be explored more thoroughly in the following sections.

AI has become an integral part of education, enabling new approaches to learning, assessment, and instructional support. Recent studies have explored various applications of AI in education, including natural language processing (NLP) for text analysis, generative AI for content creation, conversational AI for interactive learning, and adaptive AI for personalized learning experiences. These AI-driven technologies offer significant opportunities to enhance student engagement, instructional efficiency, and learning outcomes. However, they also raise concerns regarding reliability, bias, and the balance between human and AI collaboration in educational settings.

3-1-2. AI for Natural Language Processing (NLP) in Learning

Natural language processing (NLP)-based AI models have been widely applied in education for analyzing student-generated text, grading assignments, and providing formative feedback. Automated essay scoring systems, for instance, leverage large language models (LLMs) such as GPT-4 and Bard to evaluate student writing based on predefined rubrics (Yavuz, Çelik, & Yavaş Çelik, 2024). A study comparing AI-generated essay assessments with human grading found that AI tends to be more lenient on mid-to-high-performing essays but is relatively consistent in scoring lower-quality submissions. The results indicate that while AI grading models can efficiently evaluate surface-level linguistic features such as grammar and spelling, they often lack the nuanced judgment needed for content evaluation and argumentation assessment. To improve reliability, researchers recommend fine-tuning AI models based on localized rubric criteria or integrating AI feedback with human evaluation.

Beyond grading, AI is also being used for analyzing student discussion behaviors in online learning environments. Kong et al. (2025) developed an AI-powered discussion analytics dashboard that categorizes student discussion behavior based on behavioral and cognitive dimensions. Using a fine-tuned MOOC-BERT model, the system classifies student discussion posts based on participation frequency, interaction depth, and cognitive complexity. The study found that students who used AI-generated cognitive feedback demonstrated higher levels of inquiry-based learning and more structured argumentation in discussions. These findings suggest that AI-supported discussion analytics tools can help educators monitor student engagement and provide personalized feedback to enhance critical thinking skills.

Another promising application of AI in education is the automated classification of learning resources. AI models can analyze large datasets of open educational resources (OER) and categorize them based on semantic similarity and learning objectives. Li, Pardos, and Ren (2024) demonstrated how AI can align educational materials with new curricular taxonomies by using a hybrid vector-based clustering approach. This method enables AI to identify knowledge gaps, recommend relevant learning materials, and automatically translate content into multiple languages for broader accessibility. Such AI-driven classification systems have the potential to improve resource curation in LMS platforms and support personalized content delivery for diverse learners.

3-1-3. AI for Generative Content Creation in Education

Recent advancements in generative AI technologies have enabled the automated creation of instructional content, including lecture videos, slides, and study guides. AI-driven content generation tools leverage natural language processing (NLP), optical character recognition (OCR), text-to-speech (TTS), and image synthesis models to generate high-quality educational materials. Xu et al. (2024) conducted a study comparing AI-generated lecture videos with traditionally recorded instructor-led videos. The AI system automatically converted lecture slides into narrated videos using synthetic speech and facial animation technology. The findings indicated that while students demonstrated slightly higher recall scores with AI-generated content, they perceived traditional instructor-led videos as more engaging due to greater social presence. The study suggests that AI-generated instructional videos may be useful for scalable content production but should be combined with human elements to enhance student engagement.

Similarly, Netland et al. (2025) compared human-created instructional videos with AI-generated versions using tools such as ChatGPT 3.5, Midjourney, and DALL-E. Their results indicated no significant difference in overall learning outcomes between the two formats, but students rated human-created videos higher in perceived credibility and instructional clarity. Despite these findings, the study highlighted the efficiency gains of AI-generated videos, which required only 3-4 hours to produce, compared to 12-24 hours for human-produced videos. These findings reinforce the potential for AI to enhance educational content creation while underscoring the need for a balanced integration of AI and human involvement in instructional design.

3-1-4. Conversational AI for Interactive Learning

Another significant AI application in education is conversational AI, which includes chatbots and dialogue-based AI tutors designed to facilitate student engagement and scaffold learning. Recent studies have explored how AI-powered dialogue systems can support collaborative writing, legal writing instruction, and discussion-based learning. For example, CollaBot, a conversational AI agent developed by Hu, Tian, and Li (2025), was designed to support collaborative writing activities in online courses. The study found that while CollaBot did not significantly improve students' behavioral or emotional engagement, it enhanced cognitive engagement by encouraging students to explain concepts more thoroughly and structure their writing more effectively.

Similarly, AI-driven legal writing assistants have been developed to support law students in academic writing. Weber, Wambsganss, and Söllner (2024) introduced LegalWriter, a fine-tuned BERT-based AI model designed to provide real-time feedback on legal essays. The study found that students using the AI tool demonstrated significant improvements in argument structure and persuasive writing. Additionally, Chen et al. (2025) conducted a comparative study evaluating the effectiveness of AI vs. human experts in supporting reading-to-writing tasks. Their findings revealed that while AI feedback was efficient and accessible, human experts provided deeper conceptual insights, reinforcing the importance of hybrid AI-human tutoring models.

3-1-5. AI for Personalized Learning and Adaptive Feedback

AI is also being used to personalize learning experiences by adapting instructional materials and feedback to individual learner needs. AI-assisted adaptive learning systems leverage student performance data to adjust learning pathways, recommend resources, and provide targeted feedback. Darvishi et al. (2024) examined the impact of AI-generated feedback on self-regulated learning and found that AI-assisted learners exhibited greater autonomy in problem-solving tasks. However, the study also cautioned that excessive AI control can reduce student agency, emphasizing the need for a balanced approach where students can choose how and when to interact with AI support.

3-1-6. Limitations and Ethical Considerations of AI in Education

While AI technologies offer numerous benefits for education, they also present challenges related to reliability, bias, and ethical concerns. One key issue is the potential for AI grading models to introduce bias, particularly when trained on datasets that lack diverse representations of student work. Yavuz et al. (2024) noted that AI graders tend to be more lenient on high-performing students, which could lead to discrepancies in assessment fairness. Additionally, AI-generated educational content may lack the depth, emotional engagement, and instructional clarity of human-created materials (Netland et al., 2025). Ethical concerns surrounding data privacy, transparency, and AI accountability also require further examination to ensure responsible AI adoption in education.

3-2. Principles for AI adoption in education (on RQ2)

3-2-1. Understanding Education in the Context of AI Adoption

Education is a structured process of knowledge transmission, skill development, and personal growth, traditionally guided by teachers, curricula, and pedagogical frameworks. The primary goal of education is to equip learners with cognitive, social, and professional competencies that enable them to navigate the complexities of modern society. Educational theories—such as behaviorism, constructivism, and connectivism—inform teaching strategies and learning environments by emphasizing different aspects of knowledge acquisition.

With the rapid advancement of artificial intelligence (AI), educational institutions are increasingly exploring how AI can enhance teaching, learning, and assessment. AI technologies offer the potential to personalize learning, automate administrative tasks, and facilitate interactive learning experiences. However, the effective adoption of AI in education requires a clear understanding of pedagogical principles, technological capabilities, and ethical considerations. The integration of AI should align with educational values, learner diversity, and institutional goals while addressing concerns related to fairness, accessibility, and human-AI collaboration.

To formulate principles for AI adoption in education, it is essential to conduct a SWOT analysis grounded in existing literature that addresses the application of AI technologies in educational settings. This analysis provides a comprehensive view of AI's impact on education and guides the development of strategies for responsible AI implementation.

3-2-2. SWOT Analysis of AI in Education

A SWOT analysis based on recent literature reviews helps identify the benefits, limitations, potential opportunities, and risks associated with AI adoption in education.

(1) Strengths: Enhancing Learning and Efficiency

AI-powered educational systems offer several advantages that enhance both teaching efficiency and student engagement. One of the most significant benefits is personalized learning, wherein AI tailors educational content to individual learners based on their needs, styles, and progress, enabling a more adaptive and self-paced learning experience (Darvishi et al., 2024). Additionally, automated feedback and assessment tools—such as AI-assisted tutoring systems and automated essay grading—provide students with immediate, formative feedback, thereby improving learning outcomes while also reducing the workload for educators (Yavuz et al., 2024). Another key advantage is the potential for enhanced student engagement through the use of AI-driven chatbots, virtual tutors, and gamified platforms, all of which foster interactive and motivating learning environments (Hu, Tian, & Li, 2025). Furthermore, AI supports data-driven decision making by enabling the analysis of large-scale educational data to identify learning gaps, predict student performance, and recommend targeted interventions (Kong et al., 2025). Collectively, these features demonstrate the transformative potential of AI in promoting more efficient, personalized, and engaging learning experiences.

(2) Weaknesses: AI's Limitations in Education

Despite its advantages, the application of AI in education also presents several limitations that must be carefully addressed. One major concern is the lack of human emotional intelligence; AI systems are unable to replicate empathy, intuition, and the nuanced interpersonal understanding that are essential for effective mentorship and emotional support in learning environments (Chen et al., 2025). Additionally, there is a risk of over-reliance on automation, where excessive dependence on AI tools may diminish students' critical thinking and problem-solving abilities, as AI-generated responses can promote passive consumption rather than active engagement (Darvishi et al., 2024). Another critical issue is bias in AI models. Since these systems often rely on historical or imbalanced datasets, they may unintentionally perpetuate existing inequalities and produce unfair evaluations, particularly in automated grading and recommendation systems (Yavuz et al., 2024). Finally, technical barriers remain a significant obstacle, as the implementation of AI-driven educational technologies demands considerable infrastructure, financial investment, and ongoing maintenance—challenges that are particularly pronounced in resource-constrained educational settings (Xu et al., 2024). Addressing these limitations is essential to ensure that AI integration in education is not only innovative but also equitable and pedagogically sound.

(3) Opportunities: Expanding AI's Role in Education

The evolving landscape of artificial intelligence presents numerous opportunities to redefine educational practices and broaden access to learning. One promising application is the development of AI-augmented teaching assistants, which can alleviate teachers' administrative burdens by automating tasks such as grading, lesson planning, and resource generation—thereby allowing educators to dedicate more time to student engagement and personalized mentoring (Weber, Wambsganss, & Söllner, 2024). In parallel, intelligent learning analytics systems powered by AI can monitor student performance in real time, identify learners who are at risk of falling behind, and recommend data-informed interventions to improve educational outcomes (Li et al., 2024). Moreover, AI contributes to multimodal and inclusive education by offering services such as automatic translation, real-time captioning, and alternative content formats, which enhance accessibility for students with disabilities and those who are non-native speakers (Chen et al., 2025). Finally, AI is facilitating the expansion of lifelong and workplace learning, supporting flexible, on-demand education through micro-learning modules, upskilling pathways, and adaptive learning platforms tailored to the needs of dynamic labor markets (Netland et al., 2025). Together, these developments highlight the transformative potential of AI to create more efficient, inclusive, and learner-centered educational ecosystems.

(4) Threats: Ethical and Societal Concerns

The integration of AI in education also raises a range of critical concerns that require careful and proactive consideration. One such issue is the potential erosion of teacher authority; if AI tools are not thoughtfully implemented, they may inadvertently marginalize the educator's role, weakening the teacher-student relationship and fostering an over-reliance on technology (Netland et al., 2025). In addition, the widespread use of AI-powered learning systems entails significant privacy and data security risks, as these systems often collect and process large volumes of sensitive student data, making them vulnerable to breaches, unauthorized access, and misuse (Darvishi et al., 2024). There is also growing concern about job displacement in the education sector, particularly in areas such as grading, tutoring, and content development, where automation may reduce the demand for certain roles unless accompanied by policy initiatives that promote AI-human collaboration (Xu et al., 2024). Furthermore, ethical challenges in AI decision-making—such as the lack of transparency, explainability, and fairness—pose threats to trust and accountability in AI-mediated learning environments (Kong et al., 2025). Addressing these risks is essential to ensure that the adoption of AI supports, rather than undermines, the foundational values of education.

| Strengths | Weaknesses |
|---|---|
| - Personalized learning | - Lack of human emotional intelligence |
| - Automated feedback and assessment | - Over-reliance on automation |
| - Enhanced student engagement | - Bias in AI models |
| - Data-driven decision making | - Technical barriers |
| Opportunities | Threats |
| - AI-augmented teaching assistants | - Erosion of teacher authority |
| - Intelligent learning analytics | - Privacy and data security risks |
| - Multimodal and inclusive education | - Job displacement in education |
| - Lifelong learning and reskilling | - Ethical challenges in AI decision-making |

3-2-3. Principles for AI Adoption in Education

Based on the SWOT analysis of AI integration in education, the following principles are proposed to maximize educational benefits while mitigating potential risks. Each principle is directly informed by specific findings from the analysis, offering a balanced and forward-looking approach to AI adoption in educational contexts.

1. Pedagogically Aligned AI Integration

- *Strengths:* AI supports personalized and adaptive learning.
- *Weaknesses:* Risk of passive learning and reduced critical thinking.
- *Opportunities:* Alignment with instructional design frameworks.

AI tools should be integrated in accordance with established educational theories such as behaviorism, constructivism, and connectivism to ensure pedagogical coherence. Rather than replacing human instruction, AI should enhance cognitive engagement, offer scaffolding, and support active learning processes. Grounding AI in educational theory enables institutions to uphold instructional integrity while leveraging the benefits of technological innovation.

2. Human-Centered AI Design

- *Opportunities:* AI as a collaborative teaching assistant.
- *Threats:* Teacher displacement and ethical opacity in AI decisions.

AI should be designed as a collaborative partner that amplifies the capabilities of educators rather than replacing them. Systems must be transparent, explainable, and interpretable so that both students and instructors understand how decisions are made. Such design principles help address ethical concerns and foster trust in AI-mediated learning environments.

3. Balance Between AI Automation and Human Agency

- *Strengths:* Automated feedback improves efficiency.
- *Weaknesses:* Over-reliance on AI may reduce learner autonomy and critical thinking.
- *Threats:* Erosion of teacher authority.

A balanced approach is necessary, where AI-generated assessments and feedback complement—but do not override—human judgment. Students should be guided to critically assess AI-generated content, fostering media literacy and reflective thinking. Maintaining this balance preserves human agency in learning and sustains the vital teacher-student relationship.

4. Ethical and Inclusive AI Implementation

- *Weaknesses:* Bias in AI models.
- *Threats:* Ethical risks and equity concerns.
- *Opportunities:* Inclusive design for diverse learners.

Educational AI systems must undergo regular audits to detect and mitigate bias, ensuring that their use does not reinforce existing inequities. In addition, AI should be intentionally designed to support a wide range of learners—including students with disabilities, linguistic minorities, and those from marginalized communities—through features like real-time translation, accessibility support, and culturally responsive content. Such inclusive design enhances equity and access in digital education.

5. Robust Data Privacy and Security Measures

- *Threats:* Student data vulnerability and privacy breaches.

Given the extensive use of personal data in AI-powered education systems, institutions must enforce strict data governance policies. These should include secure data storage, encryption, clear data usage settings, and compliance with privacy regulations such as GDPR and FERPA. Both educators and students should be empowered to control how their data is collected, stored, and utilized, fostering digital trust and accountability.

6. Continuous Monitoring and Evaluation

- *Threats:* Long-term ethical implications and unforeseen impacts.
- *Opportunities:* Feedback-driven improvement and collaboration with researchers.

AI adoption in education should be treated as an evolving process, not a one-time implementation. Institutions must establish ongoing evaluation frameworks to monitor the effectiveness, impact, and ethical consequences of AI tools. Regular feedback from students, educators, and AI researchers should inform iterative refinements, ensuring that AI systems evolve responsibly and remain aligned with pedagogical goals and learner needs.

By explicitly grounding these principles in the SWOT analysis, educational institutions can adopt AI in a way that is both strategic and ethically informed. These principles ensure that AI is not only used to improve instructional efficiency but also to uphold core educational values, empower human agency, and promote equity and inclusion within digitally mediated learning environments.

3-3. Designing an AI-driven LMS architecture (AI-LMS; AI Learning Mate System) (on RQ3)

The AI Learning Mate System (AI-LMS) is a learning management system designed to enhance education through the integration of conversational AI technology. It enables instructors to design AI-driven interactions that are aligned with specific learning objectives, thereby supporting personalized learning experiences and fostering interactive engagement between students and AI agents. AI-LMS allows professors to customize AI interactions by setting up structured dialogue flows, defining AI roles, and integrating learning materials. This system enables automated quizzes, interactive discussions, and personalized feedback, supporting active learning engagement.

Most current approaches to integrating generative-conversational AI into education emphasize the development of educational agents through prompt engineering. These agents are designed to support student learning by engaging in conversational interactions, such as answering questions or delivering information in an interactive format. A representative example of such an AI-powered educational model is *Cogniti*, developed by the University of Sydney. This model demonstrates how educators can configure conversational AI agents using prompt engineering techniques to facilitate personalized, one-on-one interactions with students. While *Cogniti* allows instructors to embed domain-specific knowledge into the AI's responses, its core emphasis lies in enabling individualized dialogue between students and a pre-configured conversational agent.

The use of AI in this way can maximize its strengths and mitigate its weaknesses when certain factors are carefully considered. A critical aspect of integrating AI into education is ensuring that both instructors and learners remain at the center of the teaching and learning process. Importantly, educational dialogue should not be confined to one-on-one interactions between an AI agent and a single student. Instead, it should support dynamic, multi-party conversations—including triadic interactions among instructors, students, and AI agents, as well as collaborative dialogues involving multiple students simultaneously.

This study, therefore, emphasizes the need for more flexible and interactive instructional design capabilities, enabling educators to create diverse conversation flows, monitor learning progress in real time, and maintain pedagogical agency. By empowering both instructors and students to co-construct the learning experience, the proposed AI-LMS fosters a more interactive, inclusive, and pedagogically grounded integration of generative AI in education.

3-3-1. AI Conversation Design (AI-LMS; AI-Learning Mate System)

In the proposed LMS, educators can design AI learning activities that are defined by clear learning objectives, timeframes, and participation guidelines—similar to how assignments or quizzes are typically structured. However, unlike traditional assignments (which involve student submissions) or quizzes (which require responses to instructor-generated questions), AI learning activities are centered on structured, interactive conversations among instructors, students, and AI agents. These activities are designed to promote learning through dialogue-based engagement rather than one-way tasks.

| Conventional AI Tutoring Agent Model | AI-LMS (AI Learning Mate System) Model |
|---|---|
| Instructor (Sets initial rules) ↓ | (Structure and flow designed by) Instructor |
| AI Agent ←→ Learner | 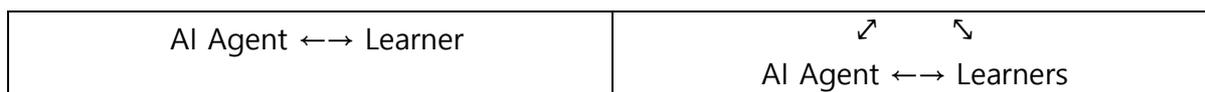 AI Agent ←→ Learners |

In this model, instructors take on an expanded role—not only facilitating learning but also acting as designers of conversational flows, configuring how AI agents engage with students to achieve specific pedagogical goals.

Instructors can customize AI-driven interactions using the following components.

| Dialogue Element | Description |
|---|---|
| AI Agent Prompt | Defines the role and behavior of the AI (e.g., tutor, feedback provider, debate partner). |
| Reference Materials | Enables the AI to draw on course materials (PDFs, articles, lecture notes) to provide contextually accurate responses. |
| Instruction for Learner | Provides learners with clear guidance on what they should do. |
| Instruction for AI | Offers the AI explicit operational instructions (e.g., how to evaluate, what feedback format to follow). |
| User Input | Accepts input from designated senders and routes it to specified recipients. A sender may be either the instructor or a learner, while recipients may include the instructor (always included), one or more learners, and/or the AI agent. |
| AI Response | Presents responses generated by the AI; visibility settings can be configured to show or hide responses from learners. |
| Templates and Placeholders | Provides reusable conversation structures and customizable placeholders for activity design. |
| Repetition (Looping) | Enables the AI to reinforce key concepts through repeated questioning or review until mastery is demonstrated. |
| Conditional Branching | Allows the conversation to follow different paths based on student input or performance, enabling differentiated and adaptive learning flows. |
| AI Team Collaboration Mode | Enables the AI to support group-based learning by facilitating discussion, coordinating team tasks, and participating in multi-party conversations with multiple learners and AI agents. |

## 3-3-2. Benefits and Limitations of AI-Driven Conversation Design

One of the key pedagogical affordances of AI-driven conversation design is its capacity to support instructor-designed chain-of-thought prompting. Unlike spontaneous reasoning generated by the AI, these conversational chains are intentionally structured by educators to guide students through sequenced thinking processes—such as recall, analysis, justification, or critique—aligned with specific instructional goals.

Using the AI-LMS, instructors can design multi-turn interactions where the AI prompts learners step-by-step, incorporating branching, repetition, and conditional scaffolding. This enables customizable reasoning flows that mirror metacognitive strategies in learning sciences. Rather than viewing prompting as a technical trick, this model positions it as a pedagogically-driven design tool under the control of the educator.

AI-driven conversation design offers a range of pedagogical benefits that enhance the quality and flexibility of teaching and learning. First, it enables customized learning experiences by allowing instructors to design AI interactions according to specific learning objectives, student needs, and instructional contexts. Through configurable roles, prompts, and content alignment, educators can create rich, adaptive learning scenarios that promote deeper engagement.

Second, AI systems provide immediate and automated feedback, supporting timely student reflection and reducing the grading and feedback workload for instructors. These interactions are structured and scalable, making it feasible to maintain responsiveness even in large courses. Additionally, AI supports reinforced learning through repetition, adaptive difficulty adjustment, and mastery-based progression, ensuring that students internalize key concepts through practice.

Third, conversation-based AI enhances student engagement by encouraging dialogue, inquiry, and critical thinking, moving learners from passive recipients of content to active participants in knowledge construction. Moreover, AI can foster collaborative and critical thinking by moderating peer discussions, prompting reflective questions, and supporting group-based learning activities, thus developing essential skills in argumentation, teamwork, and decision-making.

Despite these strengths, AI-driven conversation design also presents notable limitations. A key issue is the unpredictability and uncontrollability of generative AI outputs. Ensuring that AI-generated responses align with educational goals often requires multiple rounds of prompt engineering, testing, and fine-tuning—a process that can be time-consuming and technically demanding for instructors.

Instructors unfamiliar with AI tools or dialogue-based pedagogies may struggle to design effective learning activities, potentially leading to underutilization or pedagogical misalignment. Furthermore, the absence of emotional intelligence and contextual sensitivity in AI agents limits their ability to replicate the nuanced mentorship and affective support typically offered by human educators.

Another concern is the risk of over-reliance on AI automation, which may diminish student agency if learners become overly dependent on AI-generated solutions or feedback. In addition, issues such as algorithmic bias, data privacy, and lack of transparency in how AI generates responses remain significant ethical and technical challenges. These concerns necessitate continuous monitoring, clear usage guidelines, and robust institutional safeguards.

To mitigate these risks, institutions should invest in faculty training and instructional design support that goes beyond technical skills to include ethical awareness, prompt engineering, and pedagogical alignment. Educators must be equipped to design and manage AI interactions that support—not replace—authentic learning.

AI should be implemented within a human-centered, pedagogically grounded, and ethically responsible framework. This means educators retain instructional control, AI serves as a supportive tool, and systems are designed to ensure transparency, fairness, and student agency. When used thoughtfully, AI can amplify teaching effectiveness while preserving the irreplaceable role of human educators.

3-4. AI-LMS utilization based on each pedagogical theory (on RQ4)

The AI Learning Mate System (AI-LMS) can be adapted to different pedagogical strategies, ensuring that AI-powered learning aligns with diverse teaching methodologies. Below is how the system can be structured based on behaviorism, constructivism, and connectivism:

3-4-1. Behaviorism: Reinforcement & Repetition-Based Learning

The behaviorist theory of learning, rooted in the work of scholars such as B.F. Skinner and John B. Watson, conceptualizes learning as a process of observable behavior change driven by external stimuli and reinforcement mechanisms (Skinner, 1954; Watson, 1913;

Schunk, 2020). According to this perspective, knowledge is acquired through repetition, conditioning, and reinforcement, where correct behaviors are strengthened through positive feedback and incorrect ones are reduced through corrective responses. In educational settings, this translates into structured, linear instructional models that prioritize clear objectives, consistent practice, and immediate feedback.

The AI Learning Mate System (AI-LMS) could provide reinforcement and feedback loops by delivering immediate positive or corrective feedback in response to student input. For instance, when students answer questions correctly, the AI can offer praise or progress indicators; when errors occur, it supplies detailed explanations to reinforce the correct response. Additionally, the system supports automated repetitive learning through its looping function, allowing students to engage in repeated practice of core concepts—such as vocabulary or mathematical formulas—until mastery is achieved. The AI-LMS also facilitates conditioned learning via structured pathways, guiding learners through predefined sequences of content that gradually increase in difficulty, enabling scaffolded learning aligned with behaviorist instructional design. Finally, the integration of gamification elements—such as points, badges, and performance-based challenges—provides extrinsic motivation, reinforcing desired learning behaviors. Altogether, the AI-LMS embodies a behaviorist framework by offering structured, feedback-rich, and reward-oriented learning experiences aimed at reinforcing observable learning outcomes.

| Behaviorist Principle | AI-LMS Feature | Implementation Example |
|---|---|---|
| Reinforcement & Feedback Loops | AI provides immediate positive/negative reinforcement based on student responses. | AI rewards correct answers with encouraging messages and explains mistakes with detailed corrections. |
| Automated Repetitive Learning (Drill & Practice) | The looping function allows repeated exposure to concepts. | AI quizzes students on vocabulary or formulas until they consistently answer correctly. |
| Conditioned Learning via Structured Pathways | Predefined learning sequences guide students step by step. | AI presents increasingly difficult questions as students master earlier ones (scaffolded learning). |
| Gamification for Engagement | AI-LMS can include badges, points, and progression | Students earn points for correct answers and receive |

|  | levels. |  | performance-based challenges. |

Example: AI-Driven Quiz Activity

| No. | Element | Dialogue | Contents |
|---|---|---|---|
| 1 | AI Agent Prompt | Instructor → AI | "You are a biology professor." |
| 2 | Reference Materials | Instructor → Learner/AI | Biology course materials (PDFs) |
| 3 | Instruction for Learner | Instructor → Learner | "You will be given 10 multiple-choice questions in biology. Find the correct answers." |
| 4 | User Input | Instructor → AI | "Generate a multiple-choice question about ecological population control, without revealing the correct answer" |
| 5 | AI Response | AI → Learner | "Which factor regulates population density in an ecosystem? a) Natural disasters b) Temperature changes c) Food competition d) Rainfall" |
| 6 | User Input | Learner → Instructor | "b) Temperature changes" |
| 7 | Instruction for AI | Instructor → AI | "Student answered: *b) Temperature changes*. Give short feedback." (This instruction uses a placeholder to represent the learner's input.) |
| 8 | AI Response | AI → Learner | "The correct answer is c) Food competition. Temperature changes are considered density-independent factors, meaning they affect population size regardless of density. In contrast, food competition is density-dependent, as its impact increases with population growth." |
| 9 | Repetition |  | Repeat steps 4–8, 10 times. |

| 10 | Instruction for AI | Instructor → AI | "Give the learner final feedback for a 10-question quiz activity." |
| 11 | AI Response | AI → Learner | "Great job completing the quiz! You've demonstrated strong effort and engagement with the material. Here's a brief summary of your performance:<br>• Correct answers: 7 out of 10<br>• Most accurate topic: Population dynamics<br>• Areas to review: Ecosystem interactions and biotic/abiotic factors<br>I noticed that you're confident in identifying density-dependent factors but showed some uncertainty with environmental variables that influence population growth. To reinforce your understanding, I recommend revisiting the section on ecological relationships in your course materials." |

In this scenario, the AI-LMS can activate a looping function that delivers repeated, adaptive quiz attempts. This supports reinforcement-based learning and ensures that students internalize key concepts through practice and feedback, which aligns with behaviorist instructional principles.

3-4-2. Constructivism: Active, Inquiry-Based Learning

Constructivist learning theory, grounded in the works of Jean Piaget and Lev Vygotsky, views learning as an active, student-centered process in which learners construct knowledge through exploration, problem-solving, and reflection (Piaget, 1952; Vygotsky, 1978; Schunk, 2020). Rather than receiving knowledge passively from instructors, students engage in meaning-making through experiences, social interaction, and inquiry. This theoretical perspective emphasizes contextualized learning, scaffolding, and cognitive engagement, where learners are encouraged to question, analyze, and synthesize information based on their prior knowledge and evolving understanding.

The AI Learning Mate System (AI-LMS) embodies this constructivist approach by offering features that promote active, inquiry-based learning. First, it supports learner-centered exploration by allowing the AI to function as a guide rather than a traditional instructor. In this role, the AI prompts students with open-ended questions, encouraging them to articulate their thinking instead of simply receiving answers. Second, the system facilitates scaffolded learning through AI coaching, providing progressive hints and support as students work through complex tasks. As learners grow more confident, the AI gradually reduces the level of assistance, promoting autonomy. Third, AI-LMS enables Socratic dialogue, where the AI poses "why," "how," and "what-if" questions to stimulate reflective thinking and deeper understanding—such as engaging students in thoughtful discussions on historical or ethical topics. Lastly, the platform supports problem-based learning (PBL) by presenting real-world scenarios that require analysis and decision-making. For instance, students may work through simulated business cases with AI guidance, building problem-solving skills in authentic contexts. Together, these features align the AI-LMS with constructivist principles by fostering active engagement, critical thinking, and learner-driven discovery.

| Constructivist Principle | AI-LMS Feature | Implementation Example |
|---|---|---|
| Learner-Centered Exploration | AI serves as a guide rather than just an instructor. | AI prompts students with open-ended questions instead of giving direct answers. |
| Scaffolded Learning through AI Coaching | AI provides progressive hints to support learning. | AI gradually reduces hints as students become more confident. |
| Socratic Dialogue (Guided Discovery Learning) | AI asks why, how, and what-if questions to deepen understanding. | AI engages students in reflective discussions about historical events. |
| Problem-Based Learning (PBL) | AI presents real-world challenges for students to solve. | AI simulates business case studies, guiding students through decision-making. |

Example: AI-Facilitated Debate

| No. | Element | Dialogue | Contents |
|---|---|---|---|
| 1 | AI Agent Prompt | Instructor → AI | "You are a technology sociology expert." |
| 2 | Instruction for | Instructor → Learner | "You will participate in a debate as part of a technology sociology course. Present your arguments on the given |

| | Learner | | topic, and the AI will respond to your points during the discussion." |
|---|---|---|---|
| 3 | Instruction for AI | Instructor → AI | "We will hold a debate session for a technology sociology course. Suggest a debate topic." |
| 4 | AI Response | AI → Instructor | "Debate Topic: Should algorithmic decision-making systems be granted legal accountability in sectors like hiring, lending, and criminal justice?" |
| 5 | Instruction for Learner | Instructor → Learner | "Please provide your argument or counterargument." |
| 6 | User Input | Learner → Instructor | "I believe AI systems should be held accountable for their decisions. They are used in important areas like hiring and criminal justice. If something goes wrong, someone must take responsibility." |
| 7 | Instruction for AI | Instructor → AI | "Please respond to the following learner argument from a technology sociology expert's perspective: [Step 6 content here]." |
| 8 | AI Response | AI → Learner | "You raise an important point about accountability. As a technology sociology expert, I agree that AI systems, especially those influencing high-stakes decisions like hiring or justice, must be designed transparently and reviewed regularly. However, assigning legal accountability to non-human systems is complex—ultimately, it's the humans and institutions deploying these technologies who should bear legal responsibility and ensure ethical oversight." |
| 9 | Repetition | | Repeat steps 5–8, 5 times. |
| 10 | Instruction for learner | Instructor → Learner | "You have completed 5 rounds of debate. Please summarize each part's argument and present your final opinion." |
| 11 | Instruction for AI | Instructor → AI | "Summarize the key points discussed in the debate and provide your evaluation or reflection on the arguments presented." |

| 12 | AI Response | AI → Learner | "The debate explored whether AI systems should be legally accountable in areas like hiring and justice. One side supported accountability due to the high impact of AI decisions, while the other stressed that responsibility should remain with the humans deploying the systems. I believe legal responsibility should focus on those who design and manage AI, with strong oversight to ensure fairness and transparency." |
| 13 | User Input | Instructor → Learner | Instructor provides feedback on learner's participation and reasoning. |

Example: Virtual Counseling Simulation for Pre-Service Teachers

| No. | Element | Dialogue | Contents |
|---|---|---|---|
| 1 | Instruction for Learner | Instructor → Learner | "We will conduct a virtual counseling session with an at-risk youth as part of pre-service teacher training. These are the situation and consulting goals of today's session. As a counsellor, please carry out the session with the student." |
| 2 | Instruction for AI | Instructor → AI | "We will conduct a virtual counseling session with an at-risk youth as part of pre-service teacher training. Please provide a situation and counseling goals for this session." |
| 3 | AI Response | AI → Learner | "Situation: Minjun, a 16-year-old student, has recently shown a drop in grades, frequent absences, and social withdrawal. Teachers suspect emotional distress, possibly linked to family or peer issues. He avoids communication and often says, "I'm fine," when approached.<br>Counseling Goals: (1) Establish trust to encourage open communication. (2) Identify key stressors (academic, social, or family-related). (3) Support emotional expression and self-awareness. (4) |
| | | | Encourage small, actionable coping strategies. (5) Introduce appropriate support resources if needed. |
| 4 | AI Agent Prompt | Instructor → AI | "Please take on the role of an adolescent who is experiencing difficulties and is seeking counseling and support." |
| 5 | User Input | Learner → AI | "Hi Minjun. I'm glad you're here today. This is a safe space, and you can share whatever you're comfortable with. We're just going to talk for a bit and see how things have been going for you. There's no pressure—you can take your time. Would you like to start by telling me how you've been feeling lately?" |
| 6 | AI Response | AI → Learner | "Um… I don't really know how to start. Lately, I just feel tired all the time. School feels overwhelming, and I don't really talk to anyone much anymore. I guess I've been pretending I'm okay, but I'm not sure I really am. It's just… hard to explain." |
| 7 | Instruction for AI | Instructor → AI | "Was the goal of the counseling session achieved? Please answer with yes or no." |
| 8 | Conditional Branching | | If there is "yes", go to step 10. |
| 9 | Repetition | | Repeat steps 5–8, 10 times. |
| 10 | Instruction for AI | Instructor → AI | "Summarize and evaluate the session." |
| 11 | AI Response | AI → Learner | "Throughout the session, the counselor created a safe environment, and the student gradually opened up about emotional fatigue, academic pressure, and social withdrawal. The counselor demonstrated empathy and active listening. While no specific action plan was developed, emotional expression and trust were established. The session partially achieved its goals, especially in rapport-building and emotional awareness." |
| 12 | User Input | Instructor → Learner | Instructor provides feedback on learner's participation. |

Through this format, the AI-LMS fosters an iterative learning experience that blends guided exploration, debate, and reflection—hallmarks of constructivist pedagogy. The instructor may enable team collaboration settings and an AI rebuttal function to support structured dialogue among groups.

3-4-3. Connectivism: Networked & Collaborative Learning

Connectivism, introduced by George Siemens and Stephen Downes, is a contemporary learning theory that reflects the realities of the digital age (Siemens, 2005; Downes, 2008; Siemens, 2006). It conceptualizes learning as a networked process, where knowledge is distributed across a web of people, digital tools, and information sources. In this view, learning occurs through the ability to navigate, filter, and synthesize information from multiple nodes, emphasizing the importance of real-time access, collaboration, and adaptability. Connectivist learning values the development of digital literacy, the formation of learning networks, and the capacity to make sense of complex, ever-changing information environments.

The AI Learning Mate System (AI-LMS) supports connectivist learning by fostering interactions among students, AI agents, and diverse knowledge sources. First, it positions AI as a learning facilitator within a broader network, guiding students to external resources such as databases, academic publications, and expert forums. For example, the AI can recommend relevant research papers or open educational resources based on a student's inquiry. Second, the system encourages collaborative knowledge building by supporting team discussions and group projects, with the AI moderating forums and facilitating the co-construction of knowledge as students work together to analyze case studies. Third, AI-LMS offers dynamic, real-time learning pathways, continuously updating instructional content to reflect emerging trends and new developments. For instance, the AI can incorporate recent industry insights or policy changes into learning activities, ensuring that students engage with up-to-date information. Lastly, the system enables multi-source knowledge gathering by synthesizing content from academic articles, news, blogs, and online platforms, offering students a rich and diverse information ecosystem for deeper exploration. These features reflect the essence of connectivism by enabling learners to navigate complex knowledge networks, engage collaboratively, and construct meaning from a variety of sources.

| Connectivist Principle | AI-LMS Feature | Implementation Example |
| --- | --- | --- |

| AI as a Learning Facilitator in Networks | AI directs students to external sources, databases, and expert discussions. | AI links students to relevant research papers based on their questions. |
| Collaborative Knowledge Building | AI supports team discussions and group projects. | AI moderates online forums where students analyze case studies together. |
| Dynamic, Real-Time Learning Pathways | AI continuously updates learning paths based on new content. | AI pulls recent industry trends into discussions (e.g., latest AI developments). |
| Multi-Source Knowledge Gathering | AI synthesizes information from multiple resources. | AI integrates news articles, expert blogs, and online courses for deeper exploration. |

Example: AI-Powered Collaborative Research Project

| No. | Element | Dialogue | Contents |
| --- | --- | --- | --- |
| 1 | Instruction for Learner | Instructor → Learners | "Topic: AI and Sustainable Innovation. Explore diverse information sources—such as academic articles, news reports, and policy briefs—and submit your findings through the AI-LMS." |
| 2 | User Input | Learners → Instructor | Learners upload selected materials (e.g., PDFs, DOCs, etc.) through the AI-LMS. |
| 3 | Instruction for AI | Instructor → AI | "Summarizes, compares, and organizes the collected materials." (The files are provided) |
| 4 | AI Response | AI → Learners | "The submitted materials have been analyzed. Key Themes: AI for energy efficiency, ethical concerns, and policy approaches to sustainable innovation. Commonalities: Emphasis on AI in climate modeling and smart agriculture; shared concerns over data privacy and transparency. Differences: Diverging views on regulation—some advocate strict oversight, others prefer innovation-friendly frameworks." |

| 5 | Instruction for Learner | Instructor → Learners | "Review the synthesized summary provided by the AI Learning Mate. Based on this information, develop and submit your individual argument or position on the topic." |
| 6 | User Input | Learners → Instructor | Learners input their individual papers through the AI-LMS. |
| 7 | Instruction for AI | Instructor → AI | "Compare and synthesize the individual papers submitted by learners. Highlight key similarities and differences in their arguments." (The papers are provided) |
| 8 | AI Response | AI → Learners | "The submitted papers have been analyzed. Common Ground: Most highlight AI's role in tackling climate issues, especially in energy and agriculture. Divergent Views: Some favor strict regulation, while others support flexible, innovation-driven policies. Emerging Themes: New ideas include open-source AI and public-private partnerships for sustainability." |
| 9 | Instruction for Learner | Instructor → Learners | "Review the AI's comparative summary of the submitted papers. Reflect on the feedback and finalize your argument or perspective accordingly." |
| 10 | User Input | Learners → Instructor | Learners input their final papers through the AI-LMS. |
| 11 | User Input | Instructor → Learner | Instructor provides feedback on learner's paper. |

Example: Teaming with AI

Scenario: 3-on-3 Debate with Human-AI Teams

Overview:
A structured online debate format where AI and human students form teams (3 vs. 3). Each team supports a designated position (affirmative or negative), and teams are pre-assigned by the instructor based on their stance.

Format:
- Team Composition: Each team consists of 3 members, configured as 2 students +

- 1 AI or 3 humans vs. 3 AIs.

- Argument Source Configuration: Instructors can pre-assign whether each input is human- or AI-generated using a toggle mechanism within the AI-LMS, enabling flexible setups for hybrid or fully human/AI debates.

- Rounds: 5 rounds total.

- Length Limit: Each response or statement is limited to 120 words.

- Focus: Clear logic, supporting evidence, and persuasive clarity are prioritized.

Debate Structure Example:

1. Opening Statements – Each side presents their main argument.
2. Rebuttals – Each side responds to the opponent's claims.
3. Cross-Questions – Teams pose and respond to questions.
4. Counterarguments – Each team strengthens or clarifies its position.
5. Closing Summaries – Teams reflect and summarize their arguments.

| No. | Element | Dialogue | Contents |
|---|---|---|---|
| 1 | Instruction for Learner | Instructor → Learner | "We will conduct a 3-on-3 debate on the given topic. Teams will take either Position A or Position B. Each team member should contribute within a 120-word limit according to their assigned role." |
| 2 | Instruction for Learner | Instructor → Learners | "Team A Member 1: Please enter your Opening Statement." |
| 3-1 | User Input | Learner → Instructor | "Input an argument for the assigned role" [assigned role place holder] |
| 3-2 | AI Response | AI → Learners | "Generate an argument for the assigned role" [assigned role place holder] |
| 4 | Instruction for Learner | Instructor → Learners | "Team B Member 1: Please enter your Opening Statement." |
| 5-1 | User Input | Learner → Instructor | "Input an argument for the assigned role" [assigned role place holder] |
| 5-2 | AI Response | AI → Learners | "Generate an argument for the assigned role" [assigned role place holder] |

| 6 | Instruction for Learner | Instructor → Learners | "Team A Member 2: Please write your Rebuttal." |
| --- | --- | --- | --- |
| 7-1 | User Input | Learner → Instructor | "Input an argument for the assigned role" [assigned role place holder] |
| 7-2 | AI Response | AI → Learners | "Generate an argument for the assigned role" [assigned role place holder] |
| 8 | Instruction for Learner | Instructor → Learners | "Team B Member 2: Please write your Rebuttal." |
| 9-1 | User Input | Learner → Instructor | "Input an argument for the assigned role" [assigned role place holder] |
| 9-2 | AI Response | AI → Learners | "Generate an argument for the assigned role" [assigned role place holder] |
| 10 | Instruction for Learner | Instructor → Learners | "Team A Member 3: Please pose a Cross-Question to Team B." |
| 10-1 | User Input | Learner → Instructor | "Input an argument for the assigned role" [assigned role place holder] |
| 10-2 | AI Response | AI → Learners | "Generate an argument for the assigned role" [assigned role place holder] |
| 11 | Instruction for Learner | Instructor → Learners | "Team B Member 3: Please pose a Cross-Question to Team A." |
| 12-1 | User Input | Learner → Instructor | "Input an argument for the assigned role" [assigned role place holder] |
| 12-2 | AI Response | AI → Learners | "Generate an argument for the assigned role" [assigned role place holder] |
| 13 | Instruction for Learner | Instructor → Learners | "Team A Member 1: Please respond to the question from Team B." |
| 14-1 | User Input | Learner → Instructor | "Input an argument for the assigned role" [assigned role place holder] |
| 14-2 | AI Response | AI → Learners | "Generate an argument for the assigned role" [assigned role place holder] |
| 15 | Instruction for Learner | Instructor → Learners | "Team B Member 1: Please respond to the question from Team A." |
| 16-1 | User Input | Learner → Instructor | "Input an argument for the assigned role" [assigned role place holder] |
| 16-2 | AI | AI → | "Generate an argument for the assigned role" |
| | Response | Learners | [assigned role place holder] |
| 17 | Instruction for Learner | Instructor → Learners | "Team A Member 2: Please strengthen your argument based on the Q&A." |
| 18-1 | User Input | Learner → Instructor | "Input an argument for the assigned role" [assigned role place holder] |
| 18-2 | AI Response | AI → Learners | "Generate an argument for the assigned role" [assigned role place holder] |
| 19 | Instruction for Learner | Instructor → Learners | "Team B Member 2: Please strengthen your argument based on the Q&A." |
| 20-1 | User Input | Learner → Instructor | "Input an argument for the assigned role" [assigned role place holder] |
| 20-2 | AI Response | AI → Learners | "Generate an argument for the assigned role" [assigned role place holder] |
| 21 | Instruction for Learner | Instructor → Learners | "Team A Member 3: Please summarize your team's position and conclude the debate." |
| 22-1 | User Input | Learner → Instructor | "Input an argument for the assigned role" [assigned role place holder] |
| 22-2 | AI Response | AI → Learners | "Generate an argument for the assigned role" [assigned role place holder] |
| 23 | Instruction for Learner | Instructor → Learners | "Team B Member 3: Please summarize your team's position and conclude the debate." |
| 24-1 | User Input | Learner → Instructor | "Input an argument for the assigned role" [assigned role place holder] |
| 24-2 | AI Response | AI → Learners | "Generate an argument for the assigned role" [assigned role place holder] |
| 25 | Instruction for AI | Instructor → AI | "Evaluate the debate and provide constructive feedback for each participant based on their assigned roles." |

This approach supports the development of digital and collaborative competencies by enabling learners to interact with diverse information nodes. The AI acts as a knowledge facilitator, supporting the learner's ability to navigate and make sense of complex, evolving knowledge networks—core to connectivist learning.

## 4. Conclusion

This study presents a multi-dimensional framework for integrating generative-conversational AI into higher education through the design of an AI-powered Learning Management System (AI-LMS) and the development of pedagogically grounded instructional strategies. By examining the technological capabilities of AI, evaluating its pedagogical implications through a SWOT analysis, designing a modular LMS architecture, and aligning it with key learning theories, this research contributes a holistic and ethically informed model for AI adoption in teaching and learning environments.

The proposed AI Learning Mate System (AI-LMS) introduces practical design features—such as customizable dialogue flows, adaptive feedback, and support for collaborative learning—firmly anchored in evidence-based instructional theory. While its potential is promising, the implementation of such systems must remain informed by ongoing evaluation, user feedback, and continuous ethical reflection, particularly concerning issues of bias, data privacy, and the preservation of human agency.

In this framework, traditional LMS platforms and AI-enhanced systems like AI-LMS are envisioned as integral layers within a broader Learning Experience Platform (LXP). Whereas LMS/AI-LMS provide the structural backbone for course management, content delivery, and learner tracking, the LXP expands this foundation by integrating heterogeneous content sources, enabling personalized learning pathways, and supporting collaborative, adaptive, and self-directed learning experiences.

Within this layered ecosystem, the instructional strategies proposed in this study—categorized as drill, debate, and collaborate—serve as pedagogically grounded design patterns that align with Learning Experience Design (LXD) principles. These strategies are intended to be implemented atop the platform infrastructure to foster deep learner engagement, instructional adaptability, and meaningful AI-mediated interaction, thereby reinforcing the synergy between technology and pedagogy.

Future work will involve piloting the AI-LMS in authentic educational settings, collecting empirical data on learning outcomes and user experiences, and refining its features through iterative design. Ultimately, this research aims to contribute not only to the theoretical discourse on AI in education but also to the practical advancement of equitable, learner-centered, and pedagogically sound AI-supported learning ecosystems.